\documentclass[letterpaper, 10 pt, journal, twoside]{IEEEtran}  

\IEEEoverridecommandlockouts                              

\usepackage{amsmath} 
\usepackage{amsfonts}
\usepackage{amssymb}  
\usepackage{dsfont}
\usepackage{graphicx}
\graphicspath{{figures/}}
\usepackage{subcaption}
\usepackage{array}
\usepackage{multirow}
\usepackage{hyperref}
\usepackage{cite}
\usepackage{comment}
\usepackage[ruled,linesnumbered]{algorithm2e}
\usepackage{amsthm}
\usepackage[utf8]{inputenc}
\usepackage[english]{babel}
\usepackage{siunitx}
\usepackage{tikz}
\usetikzlibrary{positioning}
\usetikzlibrary{arrows}
\usetikzlibrary{fit}

\DeclareMathOperator{\reals}{\mathbb{R}}

\usepackage{xcolor}

\title{\Large \bf
Off The Beaten Sidewalk: Pedestrian Prediction In \\
Shared Spaces For Autonomous Vehicles
}

\author{Cyrus~Anderson$^{1}$, Ram~Vasudevan$^{2}$, and Matthew~Johnson-Roberson$^{3}$
\thanks{This work was supported by a grant from Ford Motor Company via the Ford-UM Alliance under award N022884.}
\thanks{$^1$C. Anderson is with the Robotics Institute, University of Michigan, Ann Arbor, MI 48109 USA {\tt\footnotesize andersct@umich.edu}}%
\thanks{$^2$M. Johnson-Roberson is with the Department of Naval Architecture and Marine Engineering, University of Michigan, Ann Arbor, MI 48109 USA {\tt\footnotesize mattjr@umich.edu}}%
\thanks{$^3$R. Vasudevan is with the Department of Mechanical Engineering, University of Michigan, Ann Arbor, MI 48109 USA {\tt\footnotesize ramv@umich.edu}}%
}

\begin{document}
\maketitle

\begin{abstract}
Pedestrians and drivers interact closely in a wide range of environments.
Autonomous vehicles (AVs) correspondingly face the need to predict pedestrians' future trajectories in these same environments.
Traditional model-based prediction methods have been limited to making predictions in highly structured scenes with signalized intersections, marked crosswalks, or curbs.
Deep learning methods have instead leveraged datasets to learn predictive features that generalize across scenes, at the cost of model interpretability.
This paper aims to achieve both widely applicable and interpretable predictions by proposing a risk-based attention mechanism to learn when pedestrians yield, and a model of vehicle influence to learn how yielding affects motion.
A novel probabilistic method, Off the Sidewalk Predictions (OSP), uses these to achieve accurate predictions in both shared spaces and traditional scenes.
Experiments on urban datasets demonstrate that the realtime method achieves state-of-the-art performance.

\end{abstract}


\section{Introduction}
Pedestrians and drivers interact closely in a wide range of environments.
Various road markings and signals regulate their interactions, and these features have been leveraged by many model-based prediction methods to better predict pedestrians.
Environments such as shared spaces, however, aim to regulate traffic through natural social interactions rather than traffic devices.
Shared spaces are specifically designed to minimize separation between pedestrians and drivers to promote negotiation between the two groups of road users~\cite{anvari2015modelling}.
This focus on social interactions limits the applicability of prediction methods that rely on the existence of traditional traffic devices.
Recent model-free prediction methods have instead focused on accurately predicting pedestrians in arbitrary environments.
Deep neural networks (DNNs) have proven especially effective at leveraging large datasets to learn the various interactions amongst pedestrians, between pedestrians and the environment, and between pedestrians and vehicles. 
This superior performance and generality comes at a price.
Black-box methods sacrifice both interpretability and speed with ever larger numbers of parameters.
In this work we aim to strike a balance between existing model-based and model-free methods, borrowing techniques from each. 
We introduce a probabilistic method called Off the Sidewalk Predictions (OSP) to predict pedestrian trajectories in environments where sidewalks and other traffic devices may or may not be present.
We model the pedestrian's attention similarly to the soft attention~\cite{vemula2018social} used in deep learning, and leverage existing trajectory data to learn its parameters.
At the same time, we focus on modeling only interactions between pedestrians and vehicles.
While this focus ignores interactions amongst pedestrians, we find that modeling the single type of interaction alone enables the proposed method to achieve state-of-the-art performance.
The simplified treatment of interactions also yields a model that is more interpretable and faster than state-of-the-art DNNs.
The main contributions of this work are:

\begin{figure}[t]
  \centering
  \includegraphics[width=0.45\textwidth]{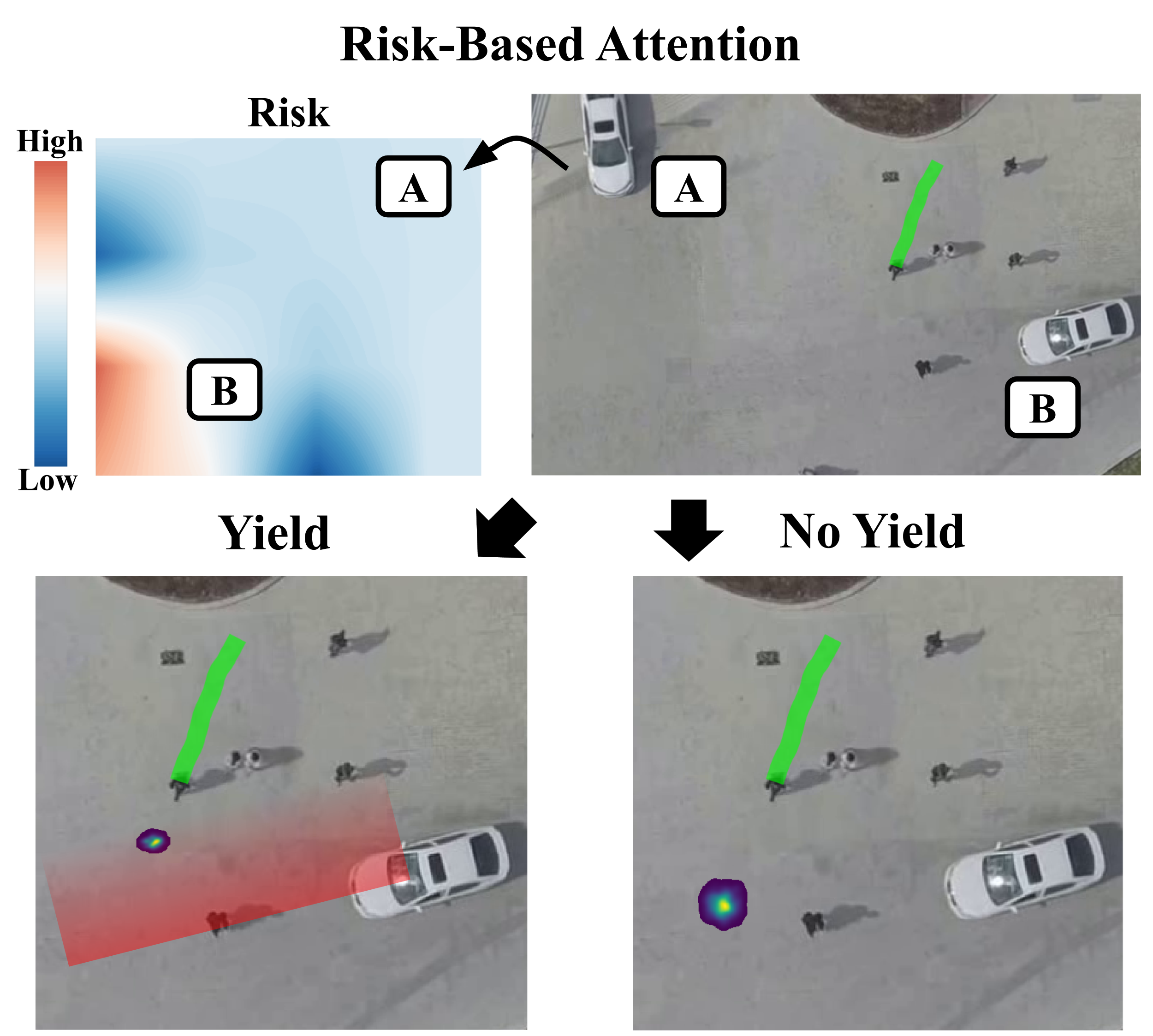}
  \caption{Steps of the proposed interaction model.
The pedestrian pays attention to each vehicle and yields in proportion to estimated risk.
A learned vehicle influence then predicts how yielding pedestrians adjust their speed, while non-yielding pedestrians continue at their desired velocity. The predicted distribution over future positions is shown for each case.}
  \label{fig:overview}
\end{figure}

\begin{enumerate}
    \item a novel and realtime probabilistic method OSP to predict pedestrian trajectories in scenes where traditional traffic devices may not be present;
    \item a tractable training procedure that avoids the auxiliary simulations or manually specified parameters called for in previous model-based works;
	\item evaluation on real-world interactions at shared spaces and urban intersections in the DUT~\cite{yang2019dut} and inD~\cite{bock2019ind} datasets.
\end{enumerate}

The proposed method OSP predicts individual pedestrians in two steps, shown in Figure~\ref{fig:overview}.
Risk-based attention is used to predict which vehicle holds the pedestrian's attention, and whether the pedestrian yields to the chosen vehicle.
To model risk in the absence of informative features such as curbs, we rely entirely on the pedestrian’s position and velocity relative to the vehicle.
For yielding pedestrians, a learned vehicle influence predicts how yielding adjusts the pedestrian's speed.
Similar to Social Forces~\cite{helbing1995sf}, vehicle influence is based on the pedestrian's distance to the vehicle's anticipated motion, but is learned from labeled data.
Since labels for attention and yielding are typically unavailable, the resulting training problem may have many modes.
We employ pseudo-likelihood techniques to decompose the problem into simpler parts that are readily solved.

The paper is organized as follows. Section~\ref{sec:related_works} describes related methods for predicting pedestrians' trajectories.
Section~\ref{sec:methods} describes the model of interactions between pedestrians and vehicles used to predict trajectories.
In Section~\ref{sec:experiments} we evaluate the model on the DUT and inD datasets, concluding in Section~\ref{sec:conclusion}.

\section{Related Work} \label{sec:related_works}

We first describe methods that predict trajectories by explicitly modeling pedestrians' interactions with other road users or the environment. 
Methods that learn models of interaction directly from large datasets are described in the next section.

\subsection{Model-Based Methods}

Recent works have had success with modeling the evolution of the pedestrian's position as a Markov process~\cite{hashimoto2015probabilistic,kooij2019context,suresh2020analysis,blaiotta2019learning}.
Methods based on solving Markov Decision Processes~\cite{kitani2012activity,karasev2016intent} and non-Markovian models such as Interacting Gaussian Processes (IGP)~\cite{trautman2010unfreezing} have also been proposed. The former, however, do not scale to account for interactions between road users.
IGP along with Social Forces based models~\cite{helbing1995sf,zeng2014sfcrosswalk,anvari2015modelling} have not achieved the same performance as more recent methods~\cite{alahi2016social}.
We adopt the Markov process approach in this paper.
In these approaches the pedestrian at each timestep chooses whether to continue a nominal trajectory or stop for an oncoming vehicle.
Previous works have modeled this decision at signaled intersections~\cite{hashimoto2015probabilistic} and marked crosswalks~\cite{blaiotta2019learning,suresh2020analysis}.
In these settings they have leveraged scene features to estimate the pedestrian's risk associated with continuing.
More general scenes containing at least curbs have been examined in Kooiji et al.~\cite{kooij2019context}, but this work addresses pedestrian motion only in one dimension.
Vehicle interactions are incorporated by measuring risk presented by the oncoming vehicle.
Measures include vehicle speed and distance~\cite{suresh2020analysis}, and minimum separation distance~\cite{kooij2019context,blaiotta2019learning}. Blaiotta~\cite{blaiotta2019learning} additionally considers the time remaining before the minimum distance is attained.
The focus of this paper is on shared scenes, which lack the informative features provided by traditional road infrastructure.
Here, risk depends only on the minimum distance and time features.

Once the decision to yield is made, many works model the pedestrian's speed as a binary option of stopping or walking~\cite{kooij2019context,blaiotta2019learning,suresh2020analysis}. 
We propose to learn a vehicle influence function that specifies how pedestrians adjust their current speed when yielding, rather than stopping.
The learned influence shown in Figure~\ref{fig:learned_weights} (right) captures the phenomenon that many pedestrians slow down before stopping. This is crucial to detecting the intent to yield early.
The training procedure proposed to learn the vehicle influence function avoids the need for auxiliary simulations~\cite{blaiotta2019learning} or manually specified parameters~\cite{helbing1995sf,zeng2014sfcrosswalk,anvari2015modelling} employed in previous works.

\subsection{Deep Learning Methods}

In contrast to traditional methods that rely heavily on manually chosen features, model-free methods learn features directly from large labeled datasets. This automatic feature selection has contributed to the recent successes of deep learning methods~\cite{lee2017desire,manh2018scene,xue2018ss}.
Unlike most model-based methods which estimate uncertainty, these initial works make only deterministic predictions.
Subsequent works have addressed this by predicting the parameters of the normal distribution~\cite{alahi2016social,chandra2019traphic}.
These works also use social pooling layers, which extract features for nearby pedestrians~\cite{alahi2016social} or road users~\cite{chandra2019traphic} based on a grid of specified size.
The fixed size of the grid, however, could fail to account for distant interactions.
Many methods have addressed this by replacing social pooling with soft attention, which models each pairwise interaction between road users~\cite{vemula2018social,sadeghian2018sophie,kosaraju2019social,ma2019trafficpredict}.
The risk-based attention used in this paper is similar to soft attention.
Of the above DNNs using soft attention, only TrafficPredict~\cite{ma2019trafficpredict} models interactions between pedestrians and vehicles.

Speed has been an area of focus for these works since the number of pairwise interactions computed for soft attention quickly grows with the number of road users.
Social pooling also entails a costly pooling step for each road user.
Social GAN~\cite{gupta2018socialgan} introduces permutation invariance to replace these slower operators.
This method reduces the computational burden to a single application of the proposed permutation invariant pooling module.
One drawback, however, is that the permutation invariant operators do not preserve the uniqueness of interaction features for each road user.
Multi-Agent Tensor Fusion (MATF)~\cite{zhao2019multi} addresses this by introducing a global pooling layer that preserves uniqueness.
Though not as efficient as Social GAN, MATF achieves state-of-the-art performance without the computational burden of soft attention.

\section{Probabilistic Trajectory Predictions} \label{sec:methods}

We formulate the problem of predicting pedestrian trajectories in Section~\ref{ssec:problem_statement}.
Section~\ref{ssec:ped_model} introduces the probabilistic method, Off the Sidewalk Predictions (OSP), used to model pedestrians' interactions with vehicles and predict pedestrian trajectories.
Estimation of the model parameters is described in Section~\ref{ssec:model_estimation}, followed by implementation details in Section~\ref{ssec:implementation}.


\subsection{Problem Statement}\label{ssec:problem_statement}

We receive noisy observations of pedestrian position and aim to predict the true position at future timesteps.
The observation at timestep $t$ of the pedestrian in the ground plane is denoted by $\hat{x}_t \in \reals^2$.
We denote the corresponding true position by $x_t \in \reals^2$. Given observations over timesteps $t=1,\dots,k$ and a final timestep of $T$, we write the prediction task as sampling future trajectories
\begin{align}
    \{x_t\}_{t=k+1}^T \sim p(\{x_t\}_{t=k+1}^T | \{\hat{x}_t\}_{t=1}^k).
\end{align}
In this work we focus on modeling interactions of a single pedestrian with multiple vehicles.
Like other model-based works~\cite{kooij2019context,blaiotta2019learning,suresh2020analysis}, we assume vehicle position and velocity for each timestep is known and deterministic.
While this assumption is not true when pedestrians and drivers repeatedly respond to each others' actions, it does hold in a scenario of significance to AVs.
In particular, the assumption holds when the AV is planning its own future trajectory, with no intent of aborting the execution before the final timestep of prediction. In this case the AV knows its own trajectory and does not modify it in response to the pedestrian's actions.
We examine this scenario in Section~\ref{ssec:av_planning}.

\begin{figure}[t]
  \centering
  \includegraphics[width=0.45\textwidth]{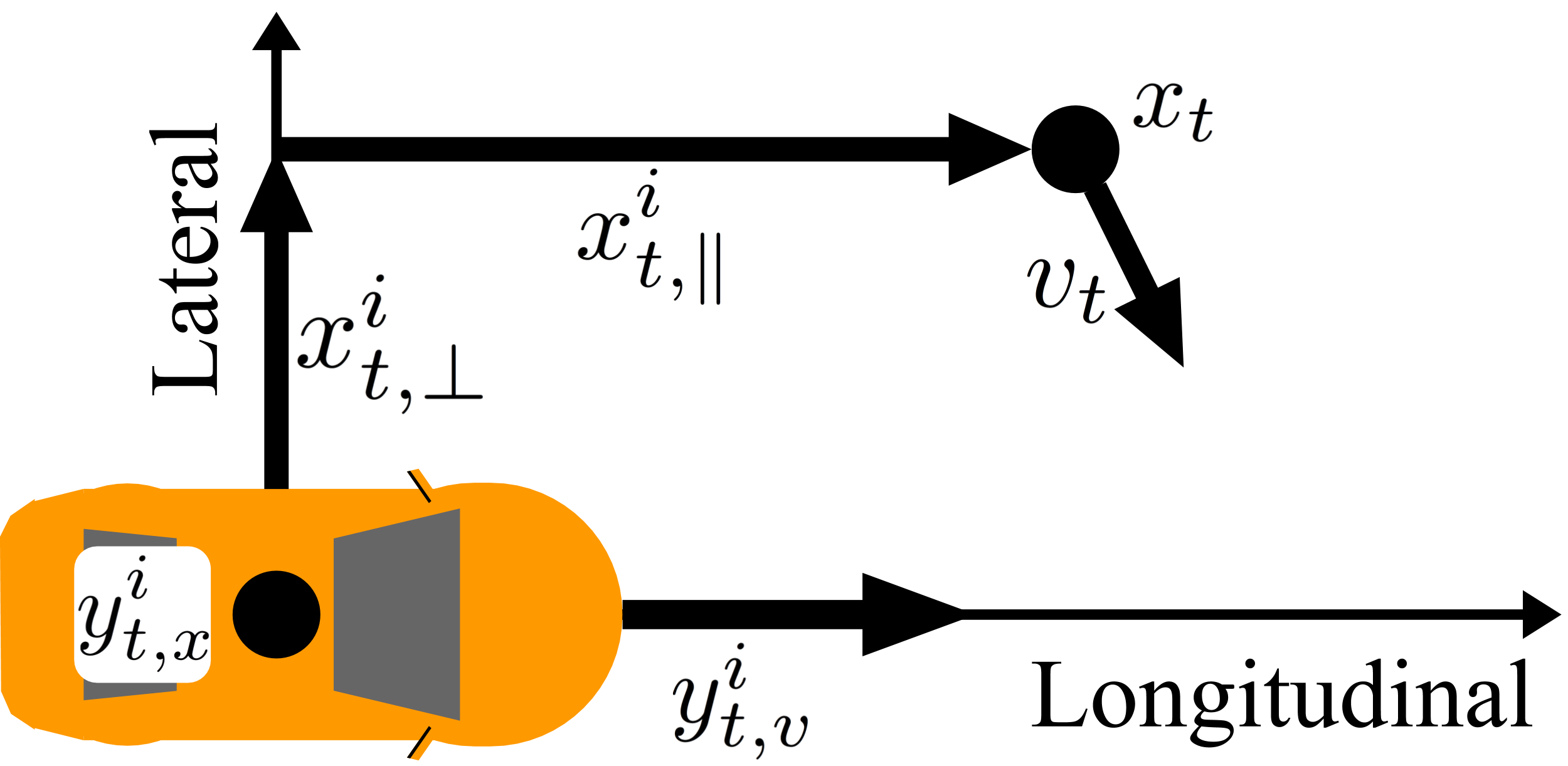}
  \caption{Reference frame for the $i$th vehicle at timestep $t$.
The pedestrian's position in this frame is decomposed into the orthogonal components $x_{t,\perp}^i$ and $x_{t,\parallel}^i$. The positions and velocities used in the world frame are shown for reference.
}
  \label{fig:vic_reference_frame}
\end{figure}
\begin{figure}[t]
  \centering
  \includegraphics[width=0.45\textwidth]{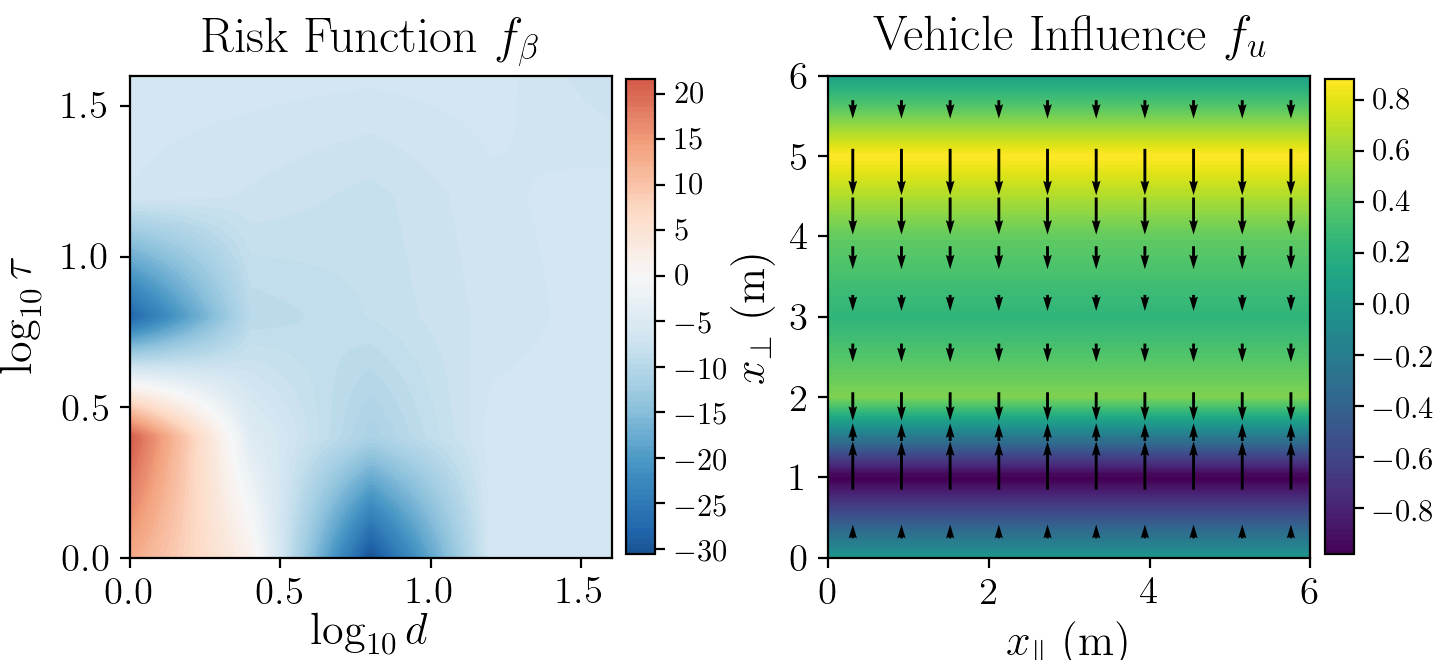}
  \caption{Learned functions for the DUT dataset. The learned risk function (left) predicts the decision boundary for pedestrians' yielding to lie along the white contour, over low values of minimum distance $d$ and time remaining $\tau$.
  The learned vehicle influence (right) resembles a curb roughly \SI{3}{\meter} away from the vehicle.
  Arrows show the movement of a yielding pedestrian with desired velocity of \SI[quotient-mode=fraction]{1}{\meter/\second}.
  }
  \label{fig:learned_weights}
\end{figure}

\subsection{Pedestrian-Vehicle Interaction Model}\label{ssec:ped_model}

Let $x_t, v_t \in \reals^2$ denote the pedestrian's position and desired velocity at timestep $t$. We now turn to defining the variables used to model the pedestrian's interaction with vehicles.
The first step is to choose the pedestrian's vehicle of focus. Let $r_t \in \{1 \dots n_v\}$ denote which of the $n_v$ vehicles currently has the pedestrian's attention.
The next step is whether or not the pedestrian yields to vehicle $r_t$. We define the binary variable $q_t = 0$ for yielding, and $q_t = 1$ for continuing at the desired velocity $v_t$.
To aid in defining the extent of interactions, we introduce $R_t \subseteq \{1,\dots, n_v\}$, the set of vehicles the pedestrian may pay attention to at timestep $t$.
Let the current position and velocity of the $i$th vehicle be given by $y_{t,x}^i,y_{t,v}^i \in \reals^2$, respectively.
Also let $x_{t,\perp}^i$ and $x_{t,\parallel}^i$ denote the lateral and longitudinal components of the pedestrian's position in the $i$th vehicle's reference frame.
This is shown in Figure~\ref{fig:vic_reference_frame}.
We define a maximum lateral distance $u_\text{max}$ to limit the extent of interactions.
Any vehicles beyond this distance are ignored by the pedestrian.
Additionally ignoring vehicles behind the pedestrian or not crossing the pedestrian's path, we define
\begin{align}
\begin{split}
    R_t = \{i \in \{1,\dots, n_v\} |
    ~ x_{t,\parallel}^i \geq -l, \\~|x_{t,\perp}^i| \leq u_{\text{max}},~
    v_t^\intercal z < 0 \},
\end{split}
\end{align}
where $l$ corresponds to half the vehicle length and $z$ corresponds to the unit vector for the lateral axis in Figure~\ref{fig:vic_reference_frame}.
The positions for which yielding may occur correspond to a subset of a quadrant in front of the vehicle, as in Figure~\ref{fig:overview}.
We now define vehicle influence over these positions.
Vehicle influence is modeled as a piecewise-linear function of $x_{t,\perp}^i$ that is symmetric about zero, with $f_{u}: \reals \rightarrow \reals$.
The function is linear in its parameters $u \in \reals^{n_u}$, with $n_u$ denoting the number of grid points.
The function grid is parameterized by its maximum distance $u_{\text{max}}$ with the $n_u$ grid points evenly spaced within $[0, u_{\text{max}}]$.
Given a pedestrian yielding to vehicle $i$, the pedestrian's velocity is defined as
\begin{align}
    v_{\text{yield}} = f_{u}(x_{t,\perp}^i) v_t.
\end{align}
The vehicle influence specifies the fraction of desired speed that is used during yielding.
Similar to Social Forces~\cite{helbing1995sf} each position specifies a yielding velocity for the pedestrian. Shown in Figure~\ref{fig:learned_weights}, the model has learned that yielding pedestrians slow down before stopping closer to the vehicle's path.
Since the pedestrian moves at their desired velocity when not yielding, we may now write the pedestrian's next position as
\begin{align}\label{eqn:x_transition}
    x_t = x_{t-1} + [q_t v_{t-1} + (1 - q_t)f_u(x_{t-1,\perp}^{r_t})] \Delta t,
\end{align}
where $\Delta t$ is the size of each timestep.
Any yielding that occurs is with respect to the vehicle being paid attention, given by $r_{t+1}$.
Now defining the distributions for each variable, we first assume normally distributed noise for the observations as
\begin{align}
    \hat{x}_t \sim N(x_t, \sigma_x^2),
\end{align}
with variance $\sigma_x^2$.
As in previous works~\cite{hashimoto2015probabilistic,blaiotta2019learning}, desired velocity is modeled as a driftless random walk with normally distributed innovations. Its transition is given by
\begin{align}
    v_t \sim N(v_{t-1}, \sigma_v^2),
\end{align}
where $\sigma_v^2$ is the variance of the innovations.
The pedestrian decisions for attention $r_t$ and yielding $q_t$ depend on risk features, which we define next.
Under a constant velocity, the remaining time before the pedestrian and vehicle $i \in R_t$ reach their minimum separation distance is given by
\begin{align}
    \tau_t^i = \frac{(x_t - y_{t,x}^i)^\intercal(y_{t,v}^i - v_t)}{\rVert y_{t,v}^i - v_t \lVert_2^2}.
\end{align}
Since $i \in R_t$, the $i$th vehicle is closing the distance to the pedestrian and this time is positive and finite.
The minimum distance itself is given by
\begin{align}
    d_t^i = (\rVert y_{t,x}^i - x_t \lVert_2^2 ~-~ (\tau_t^i)^2\rVert y_{t,v}^i - v_t \lVert_2^2)^{\frac{1}{2}}.
\end{align}
When both the remaining time $\tau_t^i$ and minimum separation distance $d_t^i$ are low, we would expect the perceived risk to be high.
On the other hand, a high value for either would suggest low risk.
We aim to learn this relationship from data with a piecewise-linear function similar to the vehicle influence.
The function is defined on a regular grid over $[b_0, b_1]^2 \subseteq \reals^2$ with $n_b^2$ evenly spaced points.
Denote the piecewise function $f_{\beta}: \reals^2 \rightarrow \reals$, which is linear in the model parameter $\beta$, which is a real vector with one element for each grid point and one bias term. This makes the total number of elements in $\beta$ equal to $n_b^2 + 1$.
We define the current risk perceived by the pedestrian as
\begin{align}
    \text{risk}_t^i = f_{\beta}(\log_{10} \tau_{t-1}^i, \log_{10} d_{t-1}^i).
\end{align}
The arguments to $f_{\beta}$ are in the log scale, to match the intuition that risk changes more rapidly nearer to collisions.  
Figure~\ref{fig:learned_weights} shows that this is reflected in the learned parameters.
Having defined risk, we now define the distributions for $r_t$ and $q_t$.
First, when there are no vehicles presenting risk, $R_t = \emptyset$. For this case we take $q_t = 1$ since no yielding will occur.
When there are possibly multiple vehicles, the pedestrian pays attention in proportion to risk.
For $i \in R_t$ we define the distribution of $r_t$ with the softmax function as  
\begin{align}
    p(r_t = i|x_{t-1},v_{t-1},\beta) = \frac{\exp{\text{risk}_t^i}}{\sum_{j \in R_t}\exp{\text{risk}_t^j}}.
\end{align}
Given vehicle $r_t$ has the pedestrian's attention, the binary decision to yield is distributed as
\begin{align}
    p(q_t = 0|x_{t-1},v_{t-1},r_t,\beta) = \frac{\exp{\text{risk}_t^{r_t}}}{1 + \exp{\text{risk}_t^{r_t}}}.
\end{align}
Compared to choosing a vehicle based on relative risk, the decision to yield is based on absolute risk.
We place weak Gaussian priors on the parameters $u$ and $\beta$ to ensure their estimation is well-posed. Let chosen scalars $\alpha_u$ and $\alpha_\beta$ denote the strength of these priors.
The negative log likelihoods are given by
\begin{align}
\begin{split}\label{eqn:priors}
    -\log p(u) = \alpha_u \lVert u \rVert_2^2 \\
    -\log p(\beta) = \alpha_\beta \lVert \beta \rVert_2^2
\end{split}
\end{align}
which correspond to zero-mean Gaussian priors, with precision proportional to $\alpha_u$ and $\alpha_\beta$.
Additionally we restrict the domain of each element of $u$ to the interval $[-1, 1]$.
This allows for the interpretation that pedestrians only decrease speed in response to vehicle influence.
Although pedestrians may temporarily increase speed while crossing in front of a fast moving vehicle, we approximate this by a lack of yielding rather than with the vehicle influence.
The next section describes how to estimate the model's unknown parameters $\sigma_v^2, u$, and $\beta$.

\subsection{Model Estimation}\label{ssec:model_estimation}

We optimize a likelihood function to estimate model parameters.
To keep quantities concise, we introduce additional notation.
For each of the variables $x_t, v_t, r_t, q_t$, let its bolded version denote the entire time series, such as $\mathbf{x} \equiv \{x_t\}_{t=1}^{t_f}$, where $t_f$ is the final timestep observed.
Additionally, let $s_t = (x_t, v_t)$.
Using the Markov structure of the model, we write the joint distribution of a single pedestrian's data
\begin{align}
\begin{split}
    L_{\text{full}}(\mathbf{x}, \mathbf{v}, \mathbf{r}, \mathbf{q}, \sigma_v^2, u, \beta) =\hspace{40mm}& \\
    = p(u)p(\beta) \prod_{t=2}^{t_f} p(\hat{x}_t|s_{t-1},r_t,q_t,u) p(q_t|s_{t-1},r_t,\beta) &  \\ p(r_t|s_{t-1},\beta) p(v_t|v_{t-1},\sigma_v^2) &
\end{split}
\end{align}
From the transition equation~\eqref{eqn:x_transition} there are many interacting terms. The decision variables $r_t$ and $q_t$ are also discrete.
These features suggest that the full likelihood $L_{\text{full}}$ may have many modes that can trap optimization procedures at poor local optima.
We instead work to separate this likelihood into commonly solved problems.  %
This is accomplished by first removing the need to estimate each $r_t$.
We first note that the set of possible vehicles $R_t$ effectively specifies $r_t$ when it is either empty or consists of a single vehicle.
Since $R_t$ depends on the pedestrian's position and desired velocity, we use the observed positions and a moving average of observed velocities over two seconds in their place to produce the estimate $\hat{R}_t$.
We remove the likelihood's dependency on $\mathbf{r}$ by ignoring data for all pedestrians having any timestep $t$ with $\lvert\hat{R}_t\rvert > 1$.
This pseudo-likelihood technique of ignoring component likelihoods will reduce the efficiency of our parameter estimates~\cite{gourieroux2017consistent}.
Using a large dataset to estimate the parameters, however, allows us to safely ignore this loss.
As each remaining $r_t$ is specified by $\hat{R}_t$, we remove the $p(r_t|x_{t-1},v_{t-1},\beta)$ term which no longer contributes any information.
Defining the set of timesteps with no candidate vehicles as $Q = \{t | \hat{R}_t = \emptyset \}$, we also have that $\forall t \in Q,  q_t = 1$.
Examining the transition equation~\eqref{eqn:x_transition}, we can simplify likelihood terms for $t \in Q$ as
\begin{align}
    p(\hat{x}_t|x_{t-1},v_{t-1},r_t,q_t=1,u) = p(\hat{x}_t|x_{t-1},v_{t-1}).
\end{align}
Rewriting the joint distribution of the included pedestrian in terms of $Q$ now yields
\begin{align}
\begin{split}
    L_{Q}(\mathbf{x}, \mathbf{v}, \mathbf{q}, \sigma_v^2, u, \beta) = \hspace{40mm} &\\
    = \prod_{t=2}^{t_f} p(v_t|v_{t-1},\sigma_v^2)  \prod_{t \in Q} p(\hat{x}_t|s_{t-1}) \hspace{20mm} &\\
    p(u)p(\beta) \prod_{t \notin Q} p(\hat{x}_t|s_{t-1},r_t,q_t,u) p(q_t|s_{t-1},r_t,\beta) &
\end{split} \label{eqn:lq}
\end{align}
The likelihood given by~\eqref{eqn:lq} has eliminated the dependency on $u$ and $\beta$ for the first two products.
In fact, the first two products form a Kalman smoothing problem.
Given noisy observations $\hat{x}_t$ for $t \in Q$, we estimate the true position $x_t$ for $t \in Q$, $v_t$ for all timesteps, and the variance $\sigma_v^2$.
Using $\hat{x}_t$ as an unbiased estimate of $x_t$ for $t \notin Q$, we are now in a position to use the remaining likelihoods.
We use these estimates in place of their unknown values to estimate the model parameters $u$ and $\beta$.
Denoting $\{q_t\}_{t \notin Q}$ by $\mathbf{q_c}$, the final likelihood we use is given by
\begin{align}
\begin{split}
    L_{c}(\mathbf{q_c}, u, \beta) = \hspace{55mm} &\\
        p(u)p(\beta) \prod_{t \notin Q} p(\hat{x}_t|s_{t-1},r_t,q_t,u) p(q_t|s_{t-1},r_t,\beta) &
\end{split} \label{eqn:lc}
\end{align}
Taking the negative log likelihood of~\eqref{eqn:lc} yields
\begin{align}
\begin{split}
    l_{c}(\mathbf{q_c}, u, \beta) = \hspace{55mm} \\
    \sum_{t \notin Q} \frac{\Delta t^2}{2\sigma_x^2} \lVert q_tv_t + (1-q_t)f_{u}(x_{t,\perp}^{r_t})v_t - \frac{\hat{x}_{t+1}-x_t}{\Delta t} \rVert_2^2 \\
    - \log p(q_t|x_{t-1},v_{t-1},r_t,\beta)
    + \alpha_u \lVert u \rVert_2^2 + \alpha_\beta \lVert \beta \rVert_2^2
\end{split} \label{eqn:nll_ubq}
\end{align}
If the decision to yield $q_t$ were known for each timestep, we would have two separate problems.
Fixing $q_t$ for $t \notin Q$ in addition to the estimated values of $x_t$ and $v_t$, only $u$ is unknown in the first summand.
Since the piecewise-linear function $f_{u}$ is linear in $u$, the first summand and the $u$ prior form a linear least squares problem for $u$ with box constraints.
There is no closed-form solution due to the restricted domain of $u$ being $[-1, 1]^{n_u} \subseteq \reals^{n_u}$, but it is readily solved by off-the-shelf linear programming solvers.
For the second summand, $\beta$ is the only unknown variable.
The piecewise-linear function $f_\beta$ is linear in $\beta$, so the second summand and the $\beta$ prior form a logistic regression for $\beta$.
Since the $q_t$ are not labeled in common datasets, we use block coordinate descent.
For blocks we use $(u, \beta)$ and $\mathbf{q_c}$.
We start with a random initial value for each unknown $q_t$ and solve the above subproblems for $u$ and $\beta$.
Fixing $u$ and $\beta$ makes the sum separable, so the optimal value of $q_i$ does not depend on that of $q_j$ for $i \neq j$. Choosing the optimal $q_t$ then consists of choosing the binary value that results in lower loss for the summand at timestep $t$.
Repeating these steps to convergence yields the final parameter estimates for $u$ and $\beta$.

\begin{table*}[t!]
\centering
\caption{Predictive performance on DUT and inD datasets. Evaluation metrics are shown as ADE/RMSE in meters. The proposed method OSP outperforms the baselines for both short-term and long-term predictions.}
\begin{tabular}{ |c|c|c|c|c|c|c|c|c|c|c| }
 \hline
    Dataset & \multicolumn{5}{c|}{DUT} & \multicolumn{5}{c|}{inD} \\
 \hline
  t (\si{\second})  & CV  & SGAN   & MATF-S & MATF  & OSP   & CV  & SGAN   & MATF-S & MATF   & OSP  \\
 \hline
   1   & 0.39/0.38 & 0.62/0.66 & 1.65/1.87 & 0.63/0.72 & 0.22/0.30 & 0.50/0.50 & 0.98/1.09 & 1.01/1.12 & 0.42/0.50 & 0.12/0.37 \\
   2   & 0.84/0.82 & 0.86/0.96 & 3.19/3.61 & 1.22/1.40 & 0.49/0.64 & 1.10/1.13 & 1.58/1.79 & 2.04/2.26 & 0.86/1.03 & 0.37/0.83 \\
   3   & 1.31/1.28 & 1.21/1.43 & 4.86/5.53 & 1.87/2.15 & 0.78/1.01 & 1.79/1.85 & 2.24/2.56 & 3.16/3.48 & 1.40/1.68 & 0.67/1.35 \\
   4   & 1.81/1.75 & 1.67/2.02 & 6.44/7.38 & 2.58/2.97 & 1.09/1.37 & 2.57/2.64 & 2.96/3.39 & 4.36/4.82 & 2.00/2.40 & 1.02/1.92 \\
   5   & 2.31/2.22 & 2.20/2.73 & 7.98/9.16 & 3.37/3.85 & 1.41/1.74 & 3.42/3.50 & 3.74/4.28 & 5.65/6.24 & 2.65/3.20 & 1.42/2.53 \\
\hline
\end{tabular}
\label{tab:traditional_test}
\end{table*}

\subsection{Implementation}\label{ssec:implementation}

We parameterize the risk function $f_\beta$ by a regular grid over $[0, 1.6]^2 \in \reals^2$ with a stride of 0.4. Since the function inputs are in the log scale, the range includes real distances up to roughly \SI{40}{\meter}.
Values outside the range are clipped to the nearest gridded point.
The vehicle influence $f_u$ is parameterized by points in $[0, 6] \in \reals$ with \SI{1}{\meter} spacing.
This sets the influence's maximum range $u_\text{max}$ to \SI{6}{\meter}.
The vehicle half length $l$ is set to \SI{2}{\meter}.
The priors on the learned functions' parameters~\eqref{eqn:priors} are set to be weak, with $\alpha_u = \frac{1}{20^2}$ and $\alpha_\beta = \frac{1}{10^2}$.
Few pedestrians are identified as yielding within the \si{5}-\SI{6}{\meter} range during the training process.
Lack of yielding in the range results in the farthest grid point of $f_u$ having its parameter estimate being pulled to the prior value of zero, shown in Figure~\ref{fig:learned_weights}.
This suggests that the maximum range of \SI{6}{\meter} for vehicle influence is sufficiently large.
The same effect appears in the risk function's parameters for the risk of vehicles that remain far from the pedestrian.
In setting the amount of observational noise, we follow the inD dataset guideline that its typical positioning error is less than \SI{0.1}{\meter}.
We thus set $\sigma_x$ to \SI{0.05}{\meter}.
For inference we find that importance sampling effectively samples the posterior distribution.
This enables us to avoid using slower particle filter steps as in other works~\cite{kooij2019context,blaiotta2019learning}.
Inference with the proposed method relies on receiving the sequence of vehicle positions to make predictions.
In most scenarios the positions are known up to the current time, but not into the future.
For this case we assume vehicles move at a constant velocity and extrapolate their future positions.

\section{Experiments} \label{sec:experiments}
We test the proposed method's ability to predict pedestrian trajectories in the DUT~\cite{yang2019dut} and inD~\cite{bock2019ind} urban datasets.
The DUT dataset contains nearly 1800 pedestrians' interactions with vehicles at two scenes. One scene is a marked crosswalk and the other is a shared space. Drivers and pedestrians in both scenes negotiate for priority of passage.
While the inD dataset contains no data collected at a shared space, it contains over 11500 road users' trajectories across four unsignalized intersections.

\subsection{Baselines}
We compare to baselines including state-of-the-art methods for pedestrian prediction based on DNNs:
\begin{itemize}
    \item \textbf{Constant Velocity (CV)} \textbf{:} The pedestrian is assumed to travel at a constant velocity.
    \item \textbf{Social GAN (SGAN)}\cite{gupta2018socialgan} \textbf{:} A GAN architecture using a permutation invariant pooling module to capture pedestrian interactions at different scales.
    \item \textbf{Multi-Agent Tensor Fusion (MATF)}\cite{zhao2019multi} \textbf{:} A GAN architecture using a global pooling layer to combine trajectory and semantic information.
    \item \textbf{Off the Sidewalk Predictions (OSP)} \textbf{:} The probabilistic interaction model introduced in Section~\ref{sec:methods}.
\end{itemize}
Each learning model, including the proposed method, is trained once on each dataset to make predictions on the unseen dataset.
Aside from CV, all of the compared methods make probabilistic predictions.
For evaluation we sample 100 trajectories from each to compare against the pedestrian's true trajectory.
The proposed method operates on observations made at \SI{10}{\Hz} while the other learning baselines operate at lower frequencies.
The observations made at \SI{10}{\Hz} are downsampled to \SI{5}{\Hz} for MATF.
SGAN is originally designed for \SI{2.5}{\Hz}, but is trained and evaluated at \SI{2}{\Hz} for the sake of comparison as in previous work~\cite{ma2019trafficpredict}.
All baselines are trained to make \SI{3}{\second} of observations and \SI{5}{\second} of predictions.
We also compare to the Multi-Agent Tensor Fusion method trained without semantic information. The method trained with semantic information is denoted MATF-S and the method without is denoted MATF.

\subsection{Evaluation Metrics}
Let $x_{i,t}$ denote the position of the $i$th pedestrian evaluated in the dataset at timestep $t$. The corresponding prediction denoted $\hat{x}_{i,t}$ is a random variable since each method is probabilistic.
Let $N$ denote the total number of pedestrians evaluated in the dataset.
We compare methods with the following metrics:
\begin{itemize}
    \item \textit{Average Distance Error (ADE):} The expected Euclidean distance between the true position and prediction, used in~\cite{blaiotta2019learning, alahi2016social, ma2019trafficpredict, gupta2018socialgan, zhao2019multi}. ADE at timestep $t$ is:
    \begin{align*}
        ADE(t) = \frac{1}{N}\sum_{i=1}^{N} \mathbb{E}[\lVert x_{i,t} - \hat{x}_{i,t}\rVert_2]
    \end{align*}
    \item \textit{Root Mean Squared Error (RMSE):} The square root of expected squared error between the true position and prediction, used in~\cite{chandra2019traphic,zhao2019multi}. RMSE at timestep $t$ is:
    \begin{align*}
        RMSE(t) = \sqrt{\frac{1}{N}\sum_{i=1}^{N} \mathbb{E}[\lVert x_{i,t} - \hat{x}_{i,t}\rVert_2^2]}
    \end{align*}
\end{itemize}
The ADE measures the distance between the mean of the predicted distribution over the pedestrian's position, and the true position.
This can also be viewed as the mean of the error distribution.
In contrast, the RMSE is a measure of the second moment of the error distribution. Larger errors thus influence the RMSE more than the ADE.

\begin{figure*}[h!]
\includegraphics[width=0.95\textwidth]{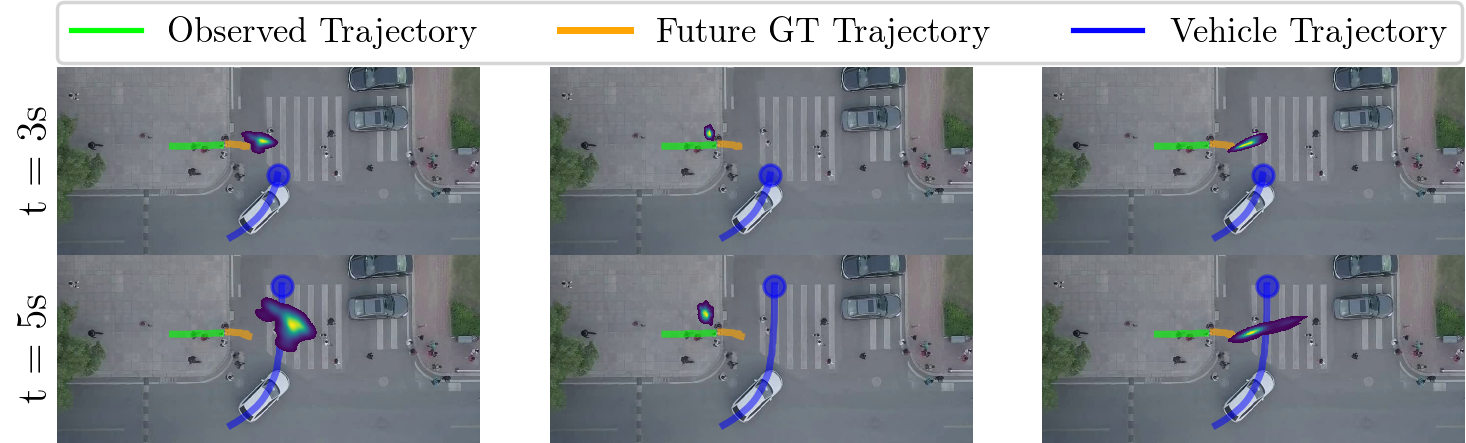}
\includegraphics[width=0.95\textwidth]{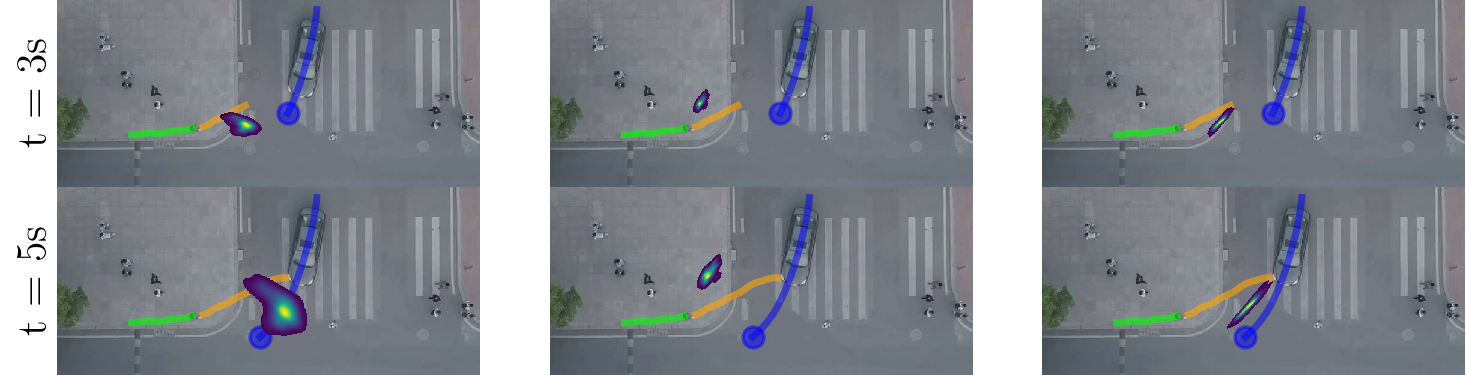}
\includegraphics[width=0.95\textwidth]{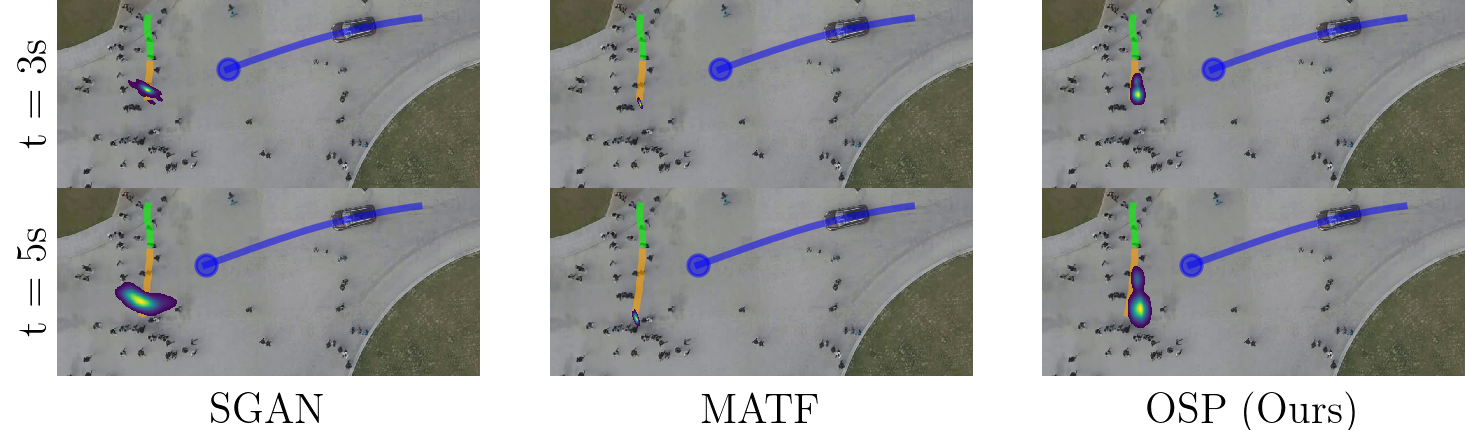}
\centering
\caption{Examples of predictions on scenes from DUT. Each method is trained on trajectory data from inD and observes \SI{3}{\second} of each road user's trajectory (pedestrian in solid green) before predicting the next \SI{5}{\second}.
Predictions for \SI{3}{\second} and \SI{5}{\second} into the future are shown for each method with likelihood according to the viridis color scale.
The orange solid line represents the ground truth trajectory of the pedestrian up to the predicted timestep.
The blue solid line represents the vehicle trajectory up to the predicted timestep.
The proposed method OSP captures uncertainty in the pedestrian's actions in both traditional crosswalk scenes and shared spaces.} \label{fig:predictions}
\end{figure*}

\subsection{Pedestrian Prediction}

Table~\ref{tab:traditional_test} shows each method's performance on both urban datasets.
The proposed method OSP outperforms previous works in both the shared spaces of DUT and unsignalized intersections of inD.
SGAN and MATF provide the next best long-term predictions for DUT and inD, respectively.
Comparing the performance of MATF-S and MATF shows that learning interactions with semantic data does not necessarily transfer from one scene to another.
Despite DUT containing a marked crosswalk scene similar to those in inD, MATF achieves better performance than MATF-S on inD without using this information.
Qualitative examples of predictions are shown in Figure~\ref{fig:predictions}.
In each example both MATF and OSP correctly predict the pedestrian's yield choice. The learned vehicle influence helps to more accurately predict where the pedestrian decides to wait.
The learned risk function also aids in capturing the uncertainty over whether the pedestrian yields.

\begin{table}[h]
\centering
\caption{Performance for simulated autonomous vehicle scenarios. Evaluation metrics are shown as ADE/RMSE in meters. Predictions made with nominal trajectory information (OSP-AV) achieve lower error than those made without (OSP).}
\begin{tabular}{ |c|c|c|c|c| }
 \hline
    Dataset & \multicolumn{2}{c|}{DUT} & \multicolumn{2}{c|}{inD} \\
 \hline
  t (\si{\second})  & OSP  & OSP-AV   & OSP  & OSP-AV  \\
 \hline
   1   & 0.23/0.30 & 0.22/0.29 & 0.28/0.38 & 0.28/0.38 \\
   2   & 0.48/0.63 & 0.47/0.61 & 0.61/0.85 & 0.60/0.84 \\
   3   & 0.74/0.99 & 0.72/0.95 & 0.99/1.37 & 0.98/1.36 \\
   4   & 1.02/1.34 & 0.98/1.30 & 1.40/1.95 & 1.39/1.93 \\
   5   & 1.29/1.69 & 1.25/1.64 & 1.85/2.57 & 1.83/2.55 \\
\hline
\end{tabular}
\label{tab:av_scenario_test}
\end{table}

\subsection{Autonomous Vehicle Planning Scenario}\label{ssec:av_planning}

Predictions inform the AV of surrounding road users' future positions.
These predictions are then used in motion planning to choose the vehicle's future trajectory.
Given that the AV knows its own nominal trajectory, it is possible to use this additional information when predicting how surrounding pedestrians interact with the AV.
We simulate this scenario in the DUT and inD datasets.
For evaluation we limit predictions to scenes containing a single moving vehicle.
The single vehicle fills the role of the AV, and its future trajectory is used alongside pedestrian observations for prediction.
Since the proposed method considers the vehicle trajectory as given, we may use the new information with no changes.
We do not compare to the baseline methods in this scenario since each is built only for making predictions based on observations up to the current time.
Performance of the proposed method using the trajectory is denoted OSP-AV and shown against the standard OSP in Table~\ref{tab:av_scenario_test}.
The trajectory information boosts the performance of OSP-AV, particularly for long-term predictions.

\subsection{Speed}

Previous sections have focused on predictive accuracy, but speed is vital to making a timely response in critical driving scenarios.
Here we benchmark the average time to make predictions for a single scenario.
Using the open source implementation of each baseline on a GTX 1080 GPU, SGAN finishes computation in \SI{0.43}{\second} and MATF in \SI{0.84}{\second}.
The average time for OSP is \SI{0.03}{\second} on a single core of an Intel Core i7-6800K CPU clocked at \SI{3.40}{\giga\Hz}.
In contrast to the deep learning methods, having fewer than 40 parameters helps to make OSP responsive.

\section{Conclusion} \label{sec:conclusion}

We propose a novel probabilistic method to predict pedestrians' trajectories in general scenes such as shared spaces, in addition to more traditional scenes.
Experiments on these scenes demonstrates that OSP achieves state-of-the-art performance.
The focus on interactions between an individual pedestrian and vehicles is both the method's strength and weakness.
The benefits include interpretable model parameters and realtime performance.
On the other hand, other types of interactions such as group interactions between pedestrians are ignored.
Modeling these types of pedestrian behaviors provides an avenue for future research.









\bibliographystyle{IEEEtran}
\bibliography{IEEEabrv,root}

\end{document}